\documentclass[letterpaper]{article} 
\usepackage[preprint]{aaai2027}  
\usepackage[hyphens]{url}  
\usepackage{graphicx} 
\usepackage{natbib}  
\usepackage{caption} 
\usepackage{algorithm}
\usepackage{algorithmic}
\usepackage{booktabs}
\usepackage{amsmath}
\usepackage{amssymb}

\newcommand{\sysname}{\textsc{AOSpec}}
\newcommand{\evd}{EVD}   
\newcommand{\insightbox}[1]{%
  \begin{center}%
  \begingroup
  \setlength{\fboxsep}{5pt}%
  \setlength{\fboxrule}{0.75pt}%
  \fcolorbox{black!80}{black!2}{%
    \parbox{0.92\columnwidth}{%
      \small\itshape
      \ensuremath{\bigstar}\enspace #1}}%
  \endgroup
  \end{center}}
\newcommand{\dtime}{D}   
\newcommand{\ttime}{T}   

\title{\sysname{}: Action and Observation Co-Speculation\\for Low-Latency Agent Serving}
\author{
    Hao (Mark) Chen, Jinnan Guo, Wayne Luk, Hongxiang Fan
}
\affiliations{
    Imperial College London, UK
}

\begin{document}

\maketitle

\begin{abstract}
Large language model agents increasingly act through stateful tools, yet model
generation and environment execution remain serialized at every step. As
decoding accelerates, tool execution becomes a growing bottleneck. Existing
action- or observation-only speculation leaves much of this latency exposed:
value is concentrated in a few slow calls, some outcomes emerge only through
execution, and longer lookahead typically requires an increasingly unlikely
chain of action predictions.
We present \sysname{}, a lossless framework that co-speculates actions and
observations across the full agent--environment loop. Expected Value Decoding
(EVD) directs observation speculation toward outcomes with the greatest
expected latency benefit, optimizing expected time hidden rather than hit
rate. For outcomes only execution can reveal, \sysname{} launches
latency-critical target actions in isolated forks that contain their effects,
while Joint Action--State Verification (JASV) verifies both the action and its
origin state against committed execution before reuse. JASV recasts
long-horizon action dependency from full-chain prediction into target
action--state verification, breaking the lookahead--accuracy tradeoff and
unlocking long-range overlap without sacrificing serial semantics.
Across Terminal-Bench serving settings spanning four harnesses, five actor
models, and five serving speeds, \sysname{} outperforms every practical
baseline, reducing mean end-to-end latency by 11.8--32.5\% and p99 latency by
up to 42.8\%. Its gains increase as decoding accelerates, and its observation
model transfers from Terminal-Bench to SWE-bench Verified without retraining.
\end{abstract}

\section{Introduction}
Modern large language model (LLM)-based agents interact with external environments through iterative cycles of reasoning, action, and observation~\citep{yao2023react}. They increasingly rely on stateful tools that inspect or modify evolving runtime environments, extending beyond simple
information retrieval. 
Frameworks such as Claude Code and OpenClaw decouple
model serving from the tool-execution environment
~\citep{anthropic2026claudecode,openclaw2026runtime}, forming what we term a disaggregated agent loop. Because the standard loop serializes model generation
and tool execution, tools account for a growing share of end-to-end latency as decoding accelerates~\citep{tilert2026, artificialanalysis_anthropic, artificialanalysis_openai, groq2024specdec, cerebras2025k2think}.

Speculation can break this serialization by starting likely future work early.
Existing systems target one side of the loop: \textit{(i)} action speculation
predicts and launches future tool calls
\citep{ye2025speculative,sui2026parallelizing,spectoolcalls2025,bpaste2026,spork2026},
whereas \textit{(ii)} observation speculation continues generation from
provisional observations verified asynchronously~\citep{saberi2026spechop}.

However, existing speculation methods fall short on heterogeneous stateful tool workloads, exposing three barriers to effective speculation.
\textbf{(1) Latency concentration:} a few slow calls dominate tool time, decoupling hit rate from latency savings (Section~\ref{sec:challenge1}).
\textbf{(2) Environment dependence:} some environment-dependent observations emerge only through execution, making observation prediction alone insufficient and requiring sandboxed action speculation (Section~\ref{sec:challenge2}).
\textbf{(3) Lookahead tradeoff:} launching speculative actions earlier increases their overlap with model generation, but lossless generation requires the intervening action chain to match;
compounding errors collapse accuracy and erase the potential speedup
(Section~\ref{sec:challenge3}). 
Together, these findings demand a new approach
to speculation across the full agent--environment loop.

\begin{figure}[!t]
\centering
\includegraphics[width=\columnwidth]{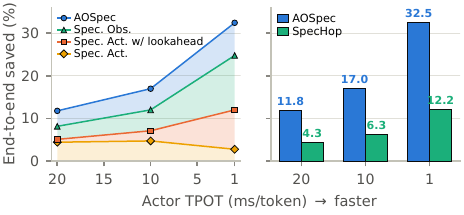}
\caption{End-to-end latency savings under trace replay, averaged over nine
(harness, model) configurations on Terminal Bench. Left: ceiling gaps among the Spec. Act, Spec. Act w/ lookahead, Spec. Obs, and
\sysname{}. Right: \sysname{} versus SpecHop at 20/10/1\,ms per token.}
\label{fig:tpot}
\end{figure}

We present \sysname{}, a lossless framework that jointly speculates actions and observations across the disaggregated agent loop
(Section~\ref{sec:method}). 
To address latency concentration, \emph{Expected
Value Decoding} (\evd) generates observation candidates by combining their probability and estimated tool time, directly optimizing expected time hidden rather than hit rate. 
For observations only execution can reveal, \sysname{} launches latency-critical target actions in isolated forks that contain their effects~\citep{cubesandbox_2026,deltabox2026}. 
To break the lookahead--accuracy tradeoff, \emph{Joint Action--State Verification} (JASV) recasts long-horizon action dependency from full-chain prediction into target action--state verification: a fork is verified and selected only when its predicted action
and origin state match the emitted action and committed state. This unlocks
long-range overlap without predicting intervening actions, while exact
verification preserves serial semantics and lossless. Together, these mechanisms yield
11.8--32.5\% end-to-end latency savings, with gains increasing as decoding
accelerates as shown in Figure~\ref{fig:tpot}.

We make three contributions:
\begin{itemize}
\item We characterize and quantify the latency concentration, information
boundary, and lookahead tension that limit speculation over stateful tools.
\item We introduce \sysname{}, combining \evd{} for latency-aware observation
drafting, isolated action execution for environment-dependent observations, and
JASV for lossless long-range lookahead without full-chain prediction.
\item Across 4 different harnesses and 5 models on Terminal-Bench, \sysname{} achieves 11.8--32.5\% end-to-end latency savings, reduces p99 latency by up to 42.8\% and outperforms practical baselines.
\end{itemize}

\section{Related Work}

\paragraph{Speculative execution for LLM agents.}
Speculative decoding accelerates generation through draft-and-verify execution~\citep{cai2024medusa, li2026eagle,chen2024hardware}. 
Agent systems extend this principle by pre-executing predicted actions or tool calls~\citep{ye2025speculative,bpaste2026,spectoolcalls2025,spork2026, zhong2026dualspec, sui2026parallelizing}, or by continuing from provisional observations~\citep{saberi2026spechop, cao2026auton}. 
Unlike prior approaches, \sysname{} jointly optimizes both forms in a lossless framework designed for stateful runtime tools with highly skewed latency, environment-dependent outputs, and side effects.

\paragraph{Agent sandboxes.}
Agent sandboxes such as CubeSandbox and DeltaBox isolate and revert file and process effects, enabling speculation over side-effecting tools \citep{cubesandbox_2026,deltabox2026}. \sysname{} uses these primitives to implement its isolation contract, while effects that cannot be isolated remain ineligible.

\section{Motivation}
\label{sec:motivation}

We model a disaggregated agent as an actor and an environment runtime. At step $i$, the actor takes $\dtime_i$ time to generate action $a_i$, and the runtime takes $\ttime_i$ time to execute it and return observation $o_i$, yielding serial latency $\sum_i(\dtime_i+\ttime_i)$. Observation speculation drafts $\hat{o}_i$ so the actor can advance while $a_i$ executes. 
Action speculation drafts and begins executing $\hat{a}_i$ before $a_i$ is emitted. 
By profiling 1{,}921 training-set trajectories across nine Terminal Bench harness--model configurations~\citep{terminalbench2},
we identify three fundamental challenges: latency concentration, environment-dependent observations, and the tension between longer overlap and compounding chain-prediction error.

\begin{figure*}[!t]
\centering
\begingroup
\newlength{\motivationpanelheight}
\setlength{\motivationpanelheight}{0.150\textwidth}
\newcommand{\motivationpanel}[3]{%
  \begin{minipage}[t]{#1}
    \centering
    \vspace{0pt}
    \parbox[t][\motivationpanelheight][t]{\linewidth}{%
      \centering
      \includegraphics[
        width=\linewidth,
        height=\motivationpanelheight,
        keepaspectratio
      ]{#2}%
    }%
    \par\vspace{0.15em}
    \footnotesize\bfseries #3
  \end{minipage}%
}
\motivationpanel{0.232\textwidth}{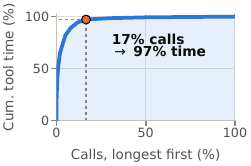}{(a) Concentration}\hfill
\motivationpanel{0.220\textwidth}{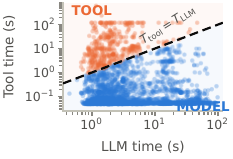}{(b) Latency scatter}\hfill
\motivationpanel{0.300\textwidth}{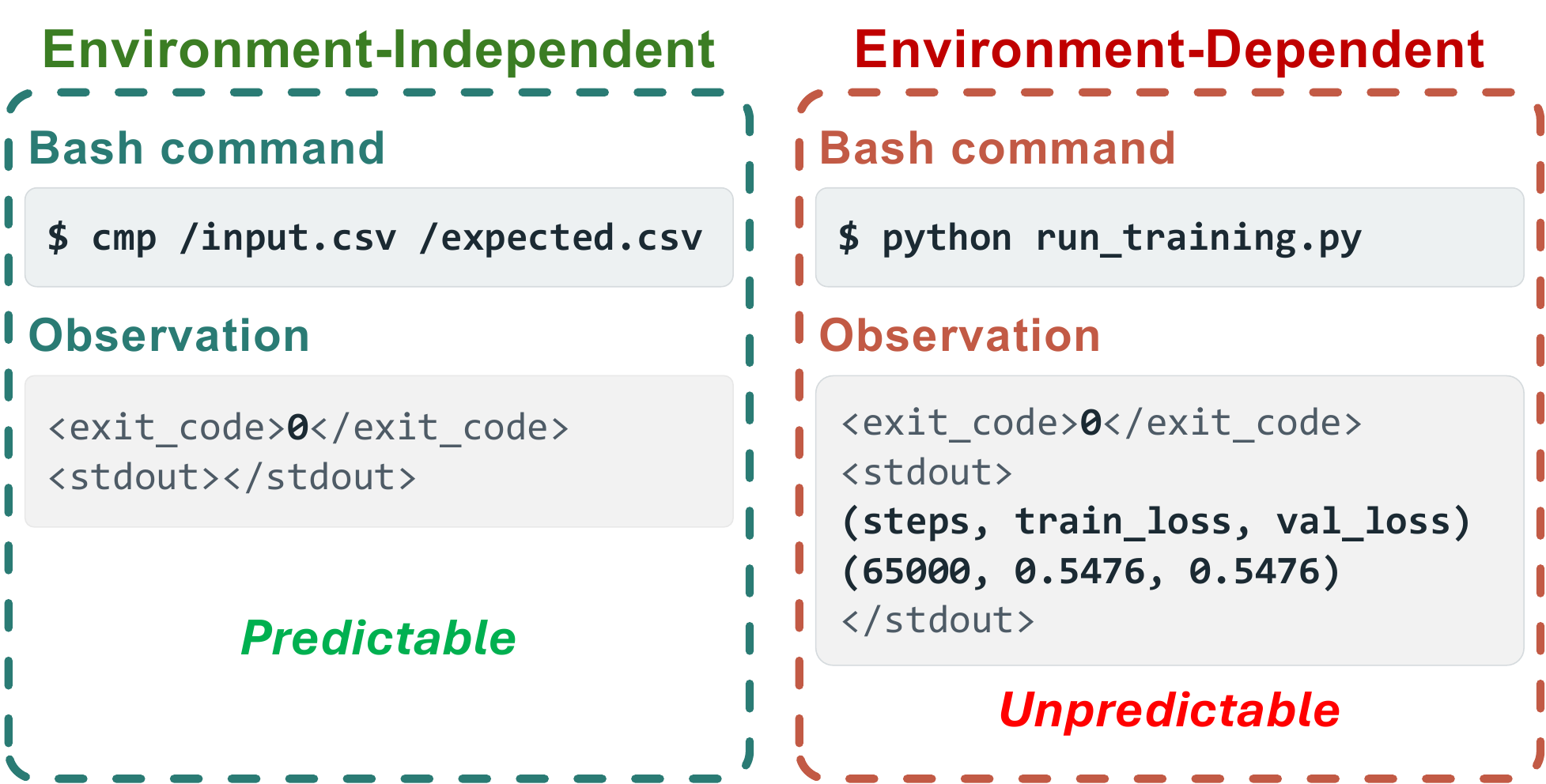}{(c) Qualitative examples}\hfill
\motivationpanel{0.233\textwidth}{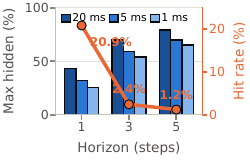}{(d) Overlap ceiling}
\endgroup
\caption{Motivating measurements from Terminal Bench.
(a) A small fraction of long calls accounts for almost all tool time.
(b) Per-step latency spans both model-bound and tool-bound regimes.
(c) Qualitative examples of costly environment-dependent observations.
(d) Longer lookahead raises the oracle overlap ceiling, but empirical exact-chain
hit rate collapses.}
\label{fig:motivation}
\end{figure*}

\subsection{Latency Concentration Decouples Hits from Savings}
\label{sec:challenge1}

As shown in Figure~\ref{fig:motivation}(a), tool time of computer-using agents is highly concentrated: the 17\% of calls lasting at least one second account for 97\% of total tool time.
Figure~\ref{fig:motivation}(b) shows the same mismatch at the step level: execution spans both model-bound and tool-bound regimes, so equal hit counts need not hide equal amounts of time.
A predictor correct only on these long-running calls would hide 97\% of tool time by achieving only 17\% hit rate. 
Conversely, a predictor correct on all remaining calls would achieve an 83\% hit rate but hide only 3\%. Thus, hit rate alone poorly reflects speculative latency savings.

\insightbox{\textbf{Insight 1:} Speculation value is concentrated: predicting
a few slow calls can hide nearly all tool time.}

\subsection{Environment Dependence Limits Observation Prediction}
\label{sec:challenge2}

Observation speculation is limited by the information available in the
actor's context. Figure~\ref{fig:motivation}(c) contrasts a predictable
file-comparison result with training metrics revealed only through execution.
More broadly, many latency-critical observations depend on hidden environment
state, such as files, running processes, or available resources, and therefore
cannot be predicted reliably from the actor's context alone.

For observations beyond this information boundary, executing a predicted
action early can supply the missing result, but only conditionally. 
Since a mispredicted action may corrupt the environment, speculative execution
can safely complement observation prediction only when contained in a sandbox
that supports snapshotting and forking.

\insightbox{\textbf{Insight 2:} Environment-dependent outcomes demand execution, not prediction: sandboxed action speculation reveals them early without risking committed state.}

\subsection{Longer Lookahead Creates Runway but Collapses Prediction Accuracy}
\label{sec:challenge3}

A speculative action launched when its issuing decode begins can hide at most
$\min(\dtime_i,\ttime_i)$. As decoding accelerates, this runway shrinks:
Figure~\ref{fig:motivation}(d) shows that the one-step overlap ceiling falls
from 43.3\% to 25.7\% as per-token latency decreases from 20\,ms to 1\,ms.

Looking farther ahead restores the lost runway: at 1\,ms, increasing the
horizon from one to five steps raises the idealized ceiling from 25.7\% to
65.2\%. 
However, prior multi-step approaches require the entire predicted action chain to match~\citep{sui2026parallelizing,bpaste2026,ye2025speculative}, causing accuracy to
collapse multiplicatively from 20.9\% at one step to 2.4\% at three and 1.2\% at five. 
The speculative timing that offers the most runway is therefore the least likely to realize it.

\insightbox{\textbf{Insight 3:} Farther lookahead creates runway, but multiplicative chain errors erase the overlap it is meant to unlock.}

Together, these findings identify three barriers to useful speculation and motivate the design of \sysname{}.

\begin{figure*}[!t]
    \centering
    \includegraphics[width=\linewidth]{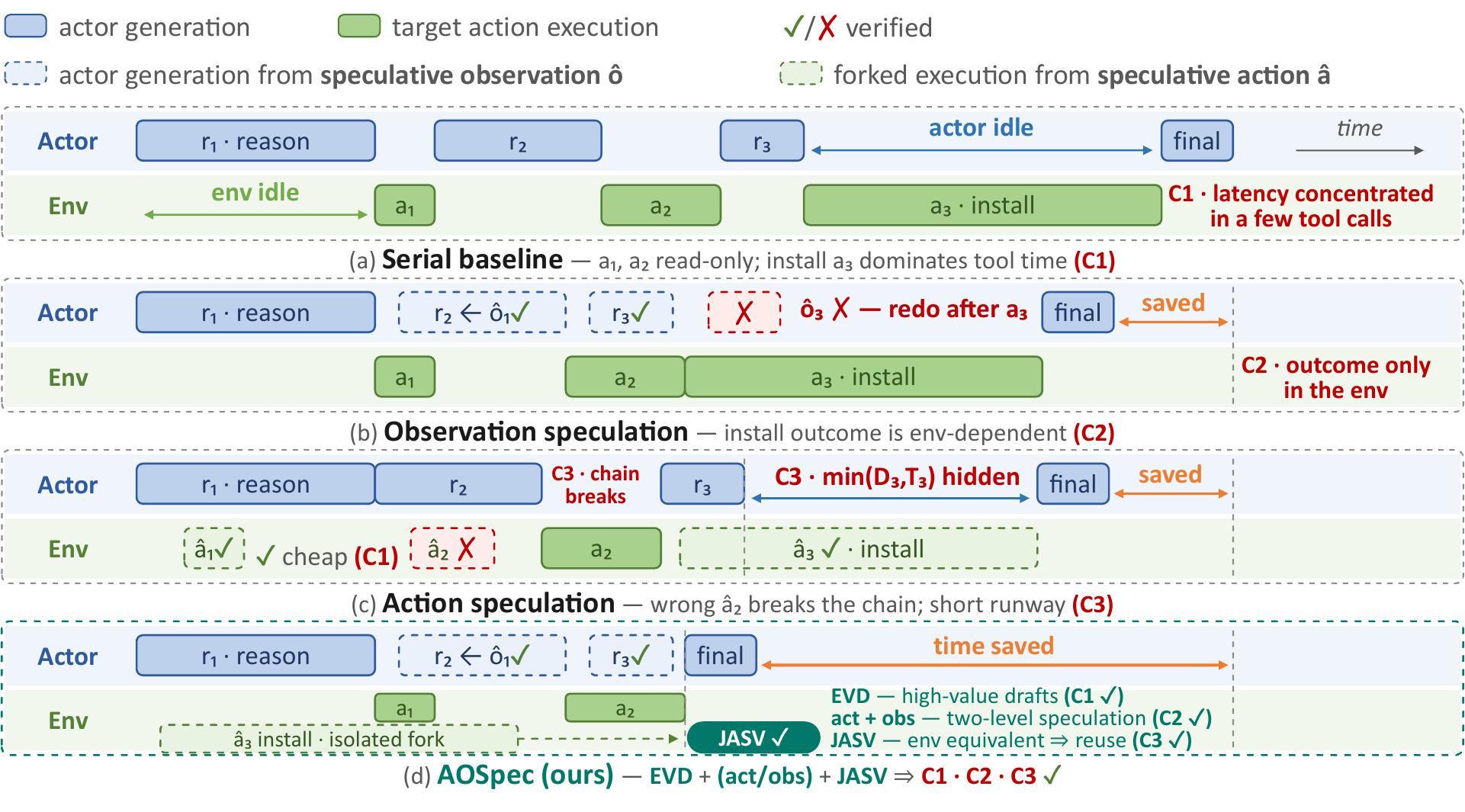}
    \caption{Overview of \sysname{} on a representative agentic trajectory. (a) Serial execution exposes tool latency. (b) Observation only speculation advances predictable outputs but cannot infer environment dependent results. (c) Action only speculation offers little same step overlap and fragile exact chain lookahead. (d) \sysname{} drafts observation with EVD and combines them with isolated target action lookahead, verifying action forks with JASV.}
    \label{fig:method-overview}
\end{figure*}

\section{Methodology}
\label{sec:method}

\subsection{Algorithm Overview}
\label{subsec:overview}

\sysname{} addresses the three requirements identified in Section~\ref{sec:motivation}:
prioritize latency over hit rate, combine predicted and environment-produced observations,
and extend action overlap without predicting an entire future trajectory.
Figure~\ref{fig:method-overview} illustrates how \sysname{} resolves the identified challenges.

At each verified observation boundary, the actor begins generating next action while speculative
actions are launched in isolated sandbox forks. Speculative candidates may target the current or a
later window. When the actor emits the target action $a$, \emph{Joint Action--State Verification} (JASV)
selects a fork only if both the speculative action matches the target action and the original environment matches the committed environment; otherwise, $a$
executes from the committed environment.

A completed action supplies its actual observation immediately. While execution remains
pending, \emph{Expected Value Decoding} (EVD) drafts candidate observations, each of which may start a provisional
actor continuation. When the actual observation returns, only the continuation whose draft is identical to the actual observation is kept. The resulting environment is then committed, and invalid forks
are pruned. Algorithm~\ref{alg:overview} summarizes the control
flow of \sysname{}.

\begin{algorithm}[t]
\caption{\sysname{}: Action--Observation Co-Speculation}
\label{alg:overview}
\small
\begin{algorithmic}[1]
\STATE $H,S\gets$ verified history and committed environment
\STATE $B\gets\textsc{StartActor}(H)$
\STATE $\mathcal F\gets\textsc{LaunchActions}(H,S)$
\WHILE{\textbf{true}}
  \STATE $x\gets\textsc{FinishActor}(B)$
  \IF{$x$ is a terminal response}
    \STATE \textbf{return} $x$
  \ENDIF
  \STATE $a\gets x$
  \STATE $f\gets$ a fork in $\mathcal F$ satisfying
         $\textsc{VerifyAction}(f,a,S)$, if one exists
  \IF{no such $f$ exists}
    \STATE $f\gets\textsc{Execute}(S,a)$
  \ENDIF

  \STATE $\mathcal P\gets\varnothing$
  \IF{$f$ has not finished}
    \STATE Start $\textsc{PredictObservations}(H,a)$
    \FOR{each candidate $\hat o$ that finishes before $f$}
      \STATE $\mathcal P[\hat o]\gets
             \textsc{StartActor}(H\circ(a,\hat o))$
    \ENDFOR
  \ENDIF
  \STATE $(o,S^+)\gets\textsc{Wait}(f)$

  \IF{$o\in\operatorname{dom}(\mathcal P)$}
    \STATE $B'\gets\mathcal P[o]$
  \ELSE
    \STATE $B'\gets\textsc{StartActor}(H\circ(a,o))$
  \ENDIF
  \STATE Cancel unfinished drafts and discard all provisional branches except $B'$

  \STATE $H\gets H\circ(a,o)$; $S\gets\textsc{Commit}(S^+)$
  \STATE $\mathcal F\gets
    \textsc{PruneInvalid}(\mathcal F\setminus\{f\},S)
    \cup\textsc{LaunchActions}(H,S)$
  \STATE $B\gets B'$
\ENDWHILE
\end{algorithmic}
\end{algorithm}

\subsection{Speculative Observation}
\label{subsec:evd}

While action execution remains pending, observation speculation advances actor
generation. Because a small fraction of calls contains most tool time
(Section~\ref{sec:challenge1}), it targets expected time hidden rather than hit rate.

\subsubsection{Optional Observation-Model Fine-Tuning}

Observation-model training could improve accuracy at smaller model sizes and candidate widths, reducing the compute required for drafting. 
Given verified history $H_t$,
action $a_t$, and observation $o_t$, we fine-tune a small model by minimizing
\begin{equation}
\mathcal{L}_{\mathrm{obs}}(\theta)
=
-\sum_{(H_t,a_t,o_t)\in\mathcal D}
\log p_\theta(o\mid H_t,a_t).
\end{equation}
We oversample costly windows to retain rare, latency-critical outcomes.
Execution traces provide supervision directly, without manual labels or
observation classes.

\subsubsection{Expected Value Decoding}

\evd{} separates observation probability from tool time. Given $(H_t,a_t)$, it constructs
a candidate set $\mathcal C(H_t,a_t)$ using one or more proposal mechanisms and estimates
each candidate's associated tool time from historical executions
$\{(o_j,\ttime_j)\}$:
\begin{equation}
\widehat T(c)
=
\frac{\sum_j K(c,o_j)\,\ttime_j}
     {\sum_j K(c,o_j)},
\end{equation}
where $K$ measures outcome similarity. Candidates are ranked by
\begin{equation}
V_o(c)
=
p_\theta(c\mid H_t,a_t)\,\widehat T(c),
\end{equation}
a proxy for expected tool time hidden. The highest-scoring candidates are drafted subject
to the branch budget. $K$ may compare discrete outcome forms or learned representations;
it affects ranking but not verification.

Observation decoding races action execution. Unfinished drafts are cancelled if execution
finishes first; a draft that finishes first starts provisional target-model generation
for step $t+1$. EVD selects what to draft, while the race determines the realized saving.

\subsubsection{Observation Verification}

When execution returns $o_t$, \sysname{} retains only a continuation whose predicted
observation is byte-identical to $o_t$. Mismatching continuations and their descendants
are discarded; if none matches, generation restarts from $o_t$. Canonicalizing
inconsequential fields such as process UUIDs could improve acceptance without changing
application-level semantics.

\subsection{Speculative Action}
\label{subsec:speculative-action}

Observation prediction cannot recover outcomes determined by the execution environment
(Section~\ref{sec:challenge2}). Speculative action instead executes speculative,
latency-critical actions in isolated forks, obtaining their actual observations while
containing their effects. JASV verifies each speculative action and its original environment
before reuse.

\subsubsection{Joint Action--State Verification}
\label{subsec:action-verification}
Each fork $f$ records its speculative action $\hat a_f$ and the version $\nu(S_f)$ of its
original environment. When the actor emits the target $a_i$ from the committed environment $S_i$,
JASV accepts the fork only if both actions and pre-execution environments match:
\begin{equation}
\mathrm{ValidAct}(f,a_i,S_i)
=
[\hat a_f=a_i]
\land
[\nu(S_f)=\nu(S_i)].
\end{equation}
Action equality covers the tool and its arguments. A valid fork executed the same action
from the same environment, so its observation and effects can be reused; otherwise, it
is discarded and $a_i$ executes from $S_i$.

JASV replaces explicit dependency analysis by discarding outstanding speculative forks upon commit. By performing speculative execution within the isolated fork, \sysname{}'s execution is semantically identical to the serial execution, making it a lossless speculation.
In Algorithm~\ref{alg:overview}, $\textsc{PruneInvalid}$ removes stale forks after each
commit, and $\textsc{VerifyAction}$ checks the action and version before reuse.

\subsubsection{State-Verified Action Lookahead}
\label{subsec:action-lookahead}

Longer lookahead increases overlap, but conventionally requires a longer chain of correct
predictions, whose acceptance probability decreases multiplicatively with the lookahead depth
(Section~\ref{sec:challenge3}). JASV instead predicts only the latency-critical target
action and later verifies its launch environment.

If an action issued at step $i$ is launched at boundary $j$, its available runway is
\begin{equation}
R_{j,i}
=
\sum_{k=j}^{i}\dtime_k
+
\sum_{k=j}^{i-1}\ttime_k,
\end{equation}
allowing it to hide up to $\min(\ttime_i,R_{j,i})$ of its execution time.

A candidate may launch at any verified boundary and remains valid across intervening
calls that leave the managed environment unchanged. Thus, these calls need not be
predicted, although forecasting a more distant target may remain harder. Calls apparent
from task context may also launch before the first actor step, similar to prefetching
before task execution.

\subsubsection{Runtime Support}

The sandboxed runtime provides snapshot, fork, execute, commit, and discard operations. Our
implementation assigns each sandbox's copy-on-write (CoW) filesystem image an immutable version, which is uniquely identified by its root hash. 
Forks record their origin version, and commit the verified speculative path, reducing JASV to a
version comparison rather than a recursive scan. Non-filesystem inputs and effects are excluded.
Our implementation adopts a CoW agent runtime from concurrent work under
anonymous review; we will add its public citation when available.

\section{Experiments}
\label{sec:experiments}

\begin{figure*}[!t]
\centering
\includegraphics[width=\textwidth]{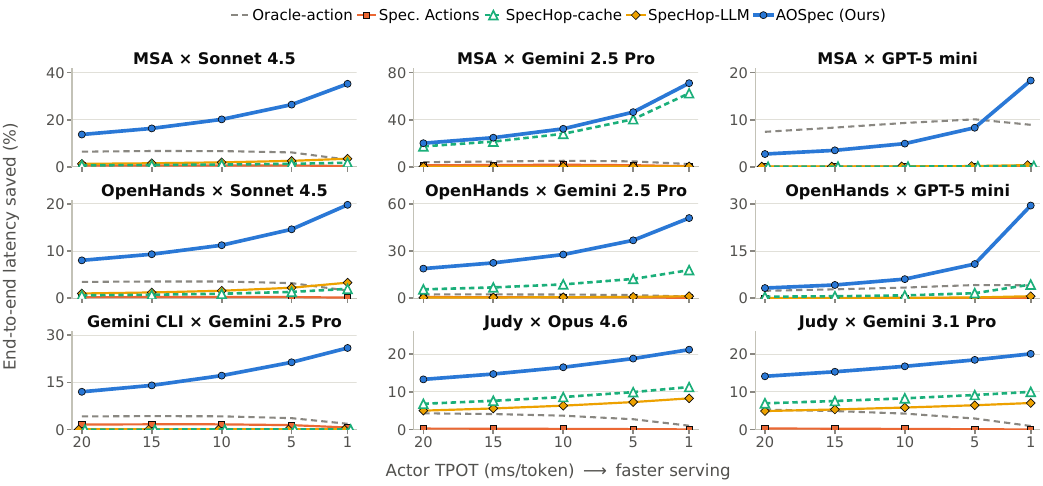}
\caption{End-to-end latency savings across nine Terminal-Bench harness and actor configurations and five actor TPOTs. Solid curves invoke an LLM for
speculation; dashed curves do not.}
\label{fig:main-results}
\end{figure*}

\begin{figure}[!t]
\centering
\includegraphics[width=\linewidth]{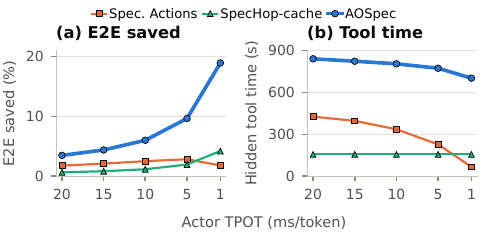}
\caption{SWE-bench transfer with the observation model trained on Terminal-Bench: (a) end-to-end latency savings and (b) total tool time hidden
across actor TPOTs.}
\label{fig:swe-generalization}
\end{figure}

\begin{figure*}[t]
\centering
\includegraphics[width=\textwidth]{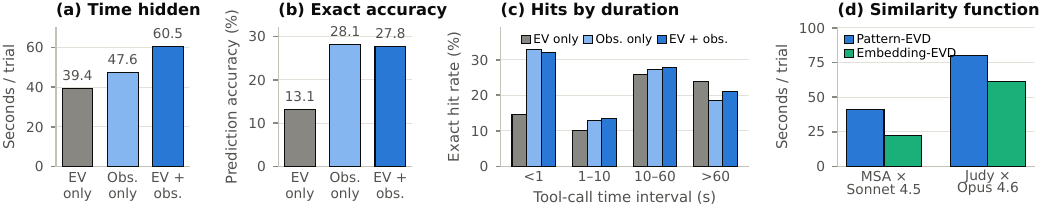}
\caption{Expected-value decoding over MSA~$\times$~Sonnet 4.5 and Judy~$\times$~Opus 4.6 (width 1, 10\,ms TPOT): (a) hidden tool time, (b) exact-match accuracy, and (c) exact-match hit rate are averaged across cells; (d) compares mined- and embedding-based value estimates for each cell.}
\label{fig:evd-ablations}
\end{figure*}

\begin{figure}[t]
\centering
\includegraphics[width=\linewidth]{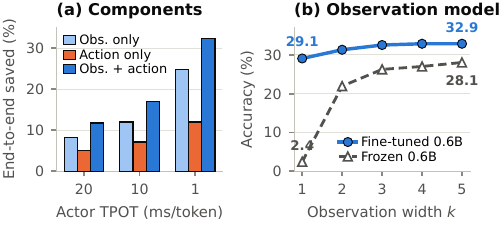}
\caption{Component and observation-model ablations. (a) Mean latency savings
from observation speculation, action speculation, and their combination across nine Terminal-Bench cells. (b) Observation accuracy for fine-tuned and frozen 0.6B models on the same MSA$\times$Sonnet trials.}
\label{fig:component-ablations}
\end{figure}

\begin{figure}[t]
\centering
\begin{minipage}{0.49\textwidth}
\centering
\includegraphics[width=\linewidth]{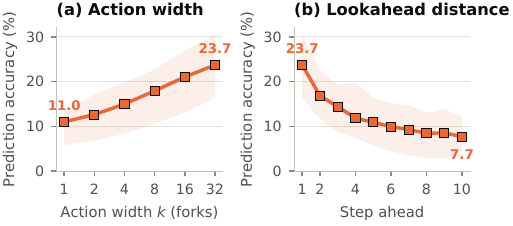}
\caption{Action-model ablations across nine Terminal-Bench cells:
trace-matched action accuracy versus (a) candidate width and (b) lookahead
distance. Curves show the nine-cell means; bands show the interquartile
ranges.}
\label{fig:action-model-ablations}
\end{minipage}
\end{figure}

\subsection{Experimental Setup}

\paragraph{Benchmarks and metrics.}
We evaluate \sysname{} through trace replay on Terminal-Bench
2.0~\citep{terminalbench2} and SWE-bench Verified~\citep{jimenez2024swebench},
spanning four agent harnesses---mini-swe-agent (MSA)~\citep{yang2024sweagent},
OpenHands~\citep{wang2025openhands}, Gemini CLI~\citep{geminicli2025}, and
Judy\footnote{Judy is an anonymous submission listed on the Terminal-Bench 2.0
leaderboard.}---and five actor models and nine harness--model configurations on
Terminal-Bench 2.0, plus MSA on SWE-bench Verified with traces from the
official ``Bash Only'' leaderboard.
We split Terminal-Bench tasks into disjoint training and test sets, ensuring
that every task category is represented in the test set. Predictors use only
training tasks.
Because the actors are accessed through APIs as black boxes
and all optimizations operate within the harness, live scheduling could cause methods to follow different trajectories.
We therefore follow the community convention to replay all methods on identical actor trajectories~\citep{xia2026idleness}, enabling reproducible comparisons across serving regimes.
We compute actor latency from recorded output token counts and sweep TPOT over
$\{20,15,10,5,1\}$\,ms/token, spanning conventional to low-latency serving
regimes~\citep{artificialanalysis_anthropic, artificialanalysis_openai, groq2024specdec, cerebras2025k2think}.
We primarily report total latency reduction relative to serial execution, alongside tool time hidden and exact prediction rate.

\paragraph{Baselines.}
We compare against Speculative Actions (\emph{Spec-Actions}), which uses
Qwen3.6-35B-A3B-FP8~\citep{qwen36_35b_a3b} to predict and launch the next
action~\citep{ye2025speculative}, and two Speculative Observation baselines, the SpecHop variants~\citep{saberi2026spechop}:
\emph{SpecHop-cache} reuses prior observations, while \emph{SpecHop-LLM} uses
Qwen3.6-35B-A3B-FP8 to draft observations. 
\emph{Oracle-action} receives the correct next action from the trace and defines the one-step action speculation ceiling. 
Serial execution is the vanilla baseline.

\paragraph{Hardware.}
Drafter generation runs on 2 NVIDIA H200 GPUs.
Tool calls are run on a Microsoft Azure Standard\_D4s\_v5 VM.
The VM has four vCPUs and 16\,GiB of memory with an Intel Xeon Platinum 8370C (Ice Lake) processor at 2.80\,GHz. Storage consists of a 64\,GB Premium SSD OS disk and a 256\,GB P15 Premium SSD data disk.

\paragraph{Implementation.}
The observation drafter is Qwen3-0.6B, fine-tuned for each harness.
The action drafter is Qwen3.6-35B-A3B-FP8 without additional training.
vLLM~\citep{kwon2023efficient} is used as the serving backend for drafting.
Unless stated otherwise, \sysname{} uses five observation branches and eight action forks.

\subsection{End-to-End Performance and Tail Latency}

\paragraph{Comparison with baselines.}
As shown in Figure~\ref{fig:main-results}, \sysname{} outperforms every
practical baseline across all 45 configuration and TPOT combinations. Its
mean saving across equally weighted configurations rises from 11.8\% at
20\,ms TPOT to 32.5\% at 1\,ms, compared with 4.3\% and 12.2\% for the
strongest baseline. Spec-Actions provides negligible savings in most
settings because long tool executions overlap only with much shorter actor
decodes. SpecHop-cache performs well when observations recur, particularly
with Gemini 2.5 Pro under MSA and OpenHands and in the Judy configurations,
but offers little benefit for Gemini CLI and most GPT-5 mini settings, where
costly outputs are less reusable. In contrast, \sysname{} remains effective
across all configurations. It also exceeds the one step action oracle in 41
of 45 settings because observation speculation and cross step action
lookahead extend beyond the one step action only ceiling. Its advantage grows
as faster decoding exposes more tool time while reducing same step overlap.

\paragraph{Tail latency.}
Table~\ref{tab:latency-distribution} reports trial-level latency at the representative 10\,ms TPOT. 
\sysname{} achieves the lowest latency at every reported quantile. It
reduces median and p90 latency by 10.7\% and 10.5\%, while the reductions grow
to 30.3\% at p95 and 42.8\% at p99. The larger tail gains agree with the
latency concentration identified in Section~\ref{sec:challenge1}.

\begin{table}[t]
\centering
\setlength{\tabcolsep}{3.0pt}
\caption{End-to-end latency distribution (seconds) per trial
at 10\,ms actor TPOT.}
\label{tab:latency-distribution}
\resizebox{\columnwidth}{!}{%
\begin{tabular}{@{}lrrrrrr@{}}
\toprule
Method & p10 & p30 & p50 & p90 & p95 & p99 \\
\midrule
Serial         & 56 & 129 & 229 & 1038 & 1842 & 7780 \\
Spec-Actions & 56 & 129 & 226 & 1036 & 1812 & 7765 \\
SpecHop-cache  & 55 & 129 & 223 & 1003 & 1624 & 5052 \\
SpecHop-LLM    & 55 & 128 & 223 & 1008 & 1779 & 7449 \\
Oracle-action  & 53 & 124 & 220 & 971  & 1669 & 7750 \\
\sysname{}     & \textbf{49} & \textbf{114} & \textbf{204} &
                 \textbf{929} & \textbf{1285} & \textbf{4449} \\
\midrule
\sysname{} reduction (\%) & 12.1 & 11.1 & 10.7 & 10.5 & 30.3 & 42.8 \\
\bottomrule
\end{tabular}
}
\end{table}

\subsection{Generalization to an Unseen Benchmark}

Figure~\ref{fig:swe-generalization} evaluates SWE-bench using the observation
model trained only on Terminal-Bench traces. At 1\,ms TPOT, \sysname{} achieves
18.9\% latency savings, compared with 4.2\% for the strongest baseline,
demonstrating transfer to an unseen benchmark without retraining.

\subsection{Expected-Value Decoding Targets Costly Calls}

\paragraph{Effect of latency weighting.}
Figure~\ref{fig:evd-ablations}(a--b) compares three decoding rules over
candidates from the same observation model: estimated tool time, model
likelihood, and their EVD product. Tool time alone favors costly but
implausible outcomes, whereas likelihood favors probable outcomes regardless
of their latency. By balancing both, EVD hides 60.5 seconds per trial, versus
47.6 seconds for likelihood and 39.4 seconds for tool time.
It redirects correct predictions toward
costly calls, raising the hit rate above 60 seconds from 18.4\% to
21.0\% while slightly reducing it for subsecond calls
(Figure~\ref{fig:evd-ablations}(c)). Thus, EVD hides more time without
requiring more correct predictions.

\paragraph{Similarity function.}
Figure~\ref{fig:evd-ablations}(d) compares two implementations of $K$:
\emph{Pattern-EVD} groups observations matched by the same mined regular
expression pattern, whereas \emph{Embedding-EVD} groups observations that are
nearby in a learned representation space. Pattern-EVD hides 40.9 versus 22.3
seconds per trial on MSA~$\times$~Sonnet 4.5 and 80.1 versus 61.2 seconds on
Judy~$\times$~Opus 4.6. We therefore use Pattern-EVD in the main evaluation.

\subsection{Ablation Studies}

\paragraph{Two-level speculation.}
Figure~\ref{fig:component-ablations}(a) shows that, at 10\,ms TPOT,
observation and action speculation alone save 12.0\% and 7.1\% on average,
respectively, while their combination saves 17.0\%. At 1\,ms TPOT, the joint
savings reach 32.5\%. The two levels are complementary: observation
speculation advances generation past predictable observations, while action
speculation executes tools early to resolve environment-dependent outputs.

\paragraph{Observation-model training.}
On the same MSA$\times$Sonnet trials, fine-tuning the 0.6B observation model
raises top-1 accuracy from 2.4\% to 29.1\%
(Figure~\ref{fig:component-ablations}(b)). 
As candidate width increases to
five, however, the frozen model reaches 28.1\% versus 32.9\% for the
fine-tuned model, narrowing the gap from 26.7 to 4.8 percentage points. 
This convergence suggests that, with sufficient compute for multiple attempts, \sysname{} can operate training-free using an off-the-shelf observation model.

\paragraph{Action-speculator accuracy.}
Increasing candidate width from 1 to 32 raises mean trace-matched accuracy
from 11.0\% to 23.7\%, trading additional forks for greater coverage
(Figure~\ref{fig:action-model-ablations}(a)). At width 32, target-action
accuracy declines from 23.7\% at distance one to 7.7\% at distance ten
(Figure~\ref{fig:action-model-ablations}(b)). By verifying the target action
and its launch state directly, JASV keeps these long-range predictions usable
without requiring the intervening action chain to match, whose accuracy is
effectively zero at this distance.

\subsection{Observation-Model Context}

\begin{table}[t]
\centering
\small
\setlength{\tabcolsep}{5pt}
\caption{Observation model input ablation on MSA~$\times$~Sonnet 4.5. Each variant removes the indicated component during both training and inference; history contains all prior reasoning, tool calls, and observations.}
\label{tab:observation-input-ablation}
\begin{tabular}{@{}lcc@{}}
\toprule
Removed input & \shortstack{Tool time\\hidden (\%)} &
\shortstack{Exact hit\\rate (\%)} \\
\midrule
None (full context)        & 27.1          & 33.3          \\
Task query                 & \textbf{28.7} & \textbf{33.6} \\
Reasoning text             & 27.3          & \textbf{33.6} \\
Prior observations         & 26.5          & 30.1          \\
All prior steps (history)  & 27.2          & 29.9          \\
\bottomrule
\end{tabular}
\end{table}

\paragraph{Effect of input components.}
As shown in Table~\ref{tab:observation-input-ablation}, removing the task query or reasoning text does not decrease both hit rate and tool time hidden.
Removing prior observations or the entire interaction history lowers exact hit rate from 33.3\% to 30.1\% and 29.9\%, respectively, while hidden time remains stable.
We therefore condition the observation model on the current action and
prior interaction history.

\section{Conclusion}

\sysname{} hides the latency of environment-execution in agentic serving by jointly
speculating observations and actions. 
EVD targets latency-critical observations
by their expected time savings, while isolated execution and JASV safely obtain environment-dependent outputs and enable long-range action speculation without predicting entire action chains. 
On representative agentic workloads, \sysname{} reduces end-to-end latency by 11.8--32.5\% and p99 latency by up to 42.8\%.
These results demonstrate that co-speculating model outputs and stateful environment execution is an effective approach to low-latency agent serving.

\bibliography{references}

\end{document}